\documentclass{vgtc}                          

\graphicspath{{figures/}{pictures/}{images/}{./}} 

\usepackage{times}                     

\usepackage{tabu}                      
\usepackage{booktabs}                  
\usepackage{lipsum}                    
\usepackage{mwe}                       

\usepackage{mathptmx}                  
\usepackage{amsmath}

\onlineid{8513}

\vgtccategory{Research}

\vgtcinsertpkg

\title{ASDF: Assembly State Detection Utilizing Late Fusion \\by Integrating 6D Pose Estimation}

\author{Hannah Schieber\thanks{e-mail: hannah.schieber@tum.de}\\ 
\and  Shiyu Li\thanks{e-mail: shiyu.li@tum.de}
\and Niklas Corell\thanks{e-mail: niklas.corell@fau.de}
\and Philipp Beckerle\thanks{e-mail: philipp.beckerle@fau.de}
\and Julian Kreimeier\thanks{e-mail: julian.kreimeier@tum.de}
\and Daniel Roth\thanks{e-mail: daniel.roth@tum.de}\\ %
\and \parbox{2in}{\scriptsize \centering Technical University of Munich  \\
Human-Centered Computing and Extended Reality Lab  \\
TUM School of Medicine and Health \\
TUM School of Computation, Information and Technology  \\ Clinic for Orthopedics and \\ Sports Orthopedics\\
TUM University Hospital, \\ Munich, Germany$^{\text{*,†,¶,‖}}$}
\and \parbox{2in}{\scriptsize \centering 
Department Artificial Intelligence in Biomedical Engineering \\ Friedrich-Alexander Universität (FAU) \\ Erlangen-Nürnberg \\
Erlangen, Germany$^{\text{*,‡,§}}$}
\and \parbox{2in}{\scriptsize \centering Chair of Autonomous Systems and Mechatronics \\
Friedrich-Alexander Universität (FAU) \\ Erlangen-Nürnberg \\
Erlangen, Germany$^{\text{§}}$}
}

\teaser{
  \centering
  \includegraphics[width=0.9\textwidth]{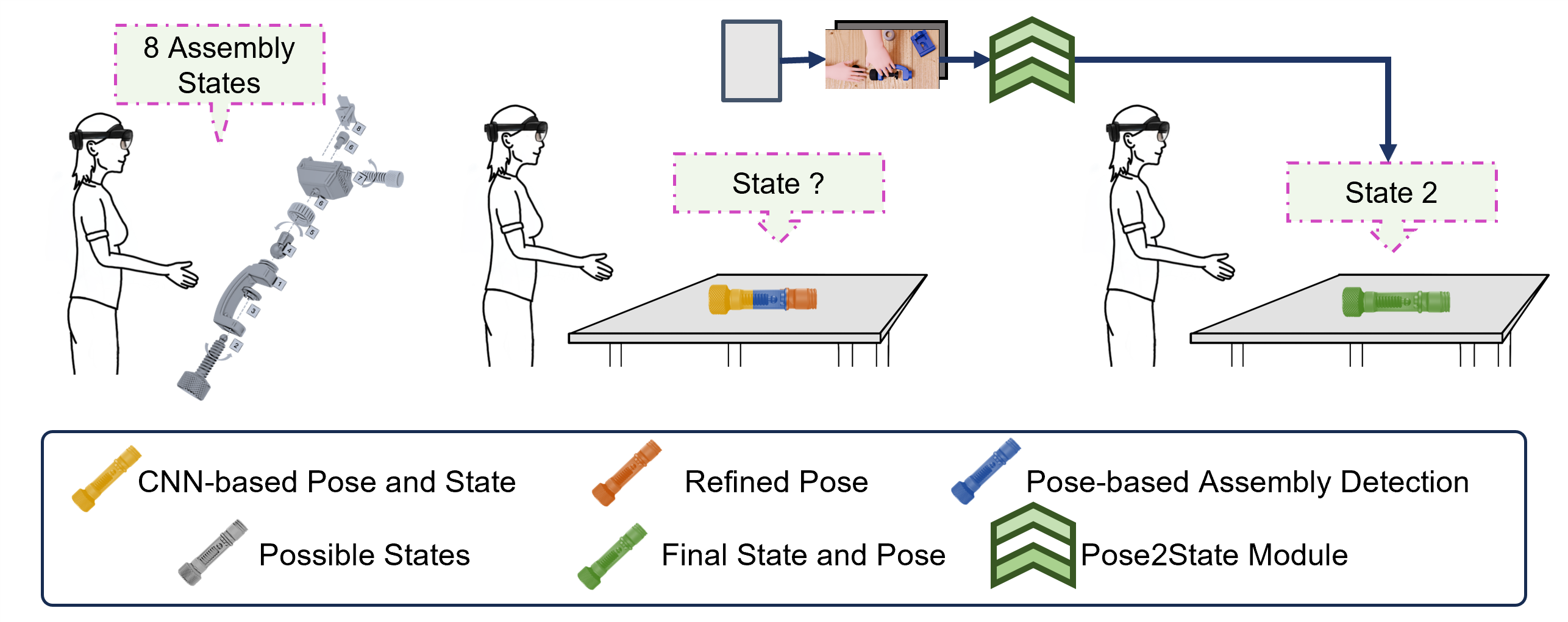}
  \caption{\textbf{We present \acf{asdf}.} Our approach fuses RGB and depth information as well as 6D pose estimation and state predictions. In our Pose2State module we calculate the final state in a late fusion step by combining 6D pose estimation and assembly state detection.  \label{fig:teaser}}

}

\abstract{
    In medical and industrial domains, providing guidance for assembly processes can be critical to ensure efficiency and safety. Errors in assembly can lead to significant consequences such as extended surgery times and prolonged manufacturing or maintenance times in industry. Assembly scenarios can benefit from in-situ augmented reality visualization, i.e., augmentations in close proximity to the target object, to provide guidance, reduce assembly times, and minimize errors. In order to enable in-situ visualization, 6D pose estimation can be leveraged to identify the correct location for an augmentation. Existing 6D pose estimation techniques primarily focus on individual objects and static captures. However, assembly scenarios have various dynamics, including occlusion during assembly and dynamics in the appearance of assembly objects. Existing work focus either on object detection combined with state detection, or focus purely on the pose estimation. To address the challenges of 6D pose estimation in combination with assembly state detection, our approach ASDF builds upon the strengths of YOLOv8, a real-time capable object detection framework. We extend this framework, refine the object pose, and fuse pose knowledge with network-detected pose information. Utilizing our late fusion in our Pose2State module results in refined 6D pose estimation and assembly state detection. By combining both pose and state information, our Pose2State module predicts the final assembly state with precision. The evaluation of our ASDF dataset shows that our Pose2State module leads to an improved assembly state detection and that the improvement of the assembly state further leads to a more robust 6D pose estimation. Moreover, on the GBOT dataset, we outperform the pure deep learning-based network and even outperform the hybrid and pure tracking-based approaches.
} 


\usepackage[nolist]{acronym}
\usepackage{multirow}
\usepackage{multicol}

\usepackage{tikz}
\begin{document}
\begin{acronym}[Bspwwww.]  

\acro{ar}[AR]{augmented reality}
\acro{asdf}[ASDF]{assembly state detection utilizing late fusion}
\acro{add}[ADD]{average distance error}
\acro{ate}[ATE]{absolute trajectory error}
\acro{bvip}[BVIP]{blind or visually impaired people}
\acro{cnn}[CNN]{convolutional neural network}
\acro{fov}[FoV]{field of view}
\acro{fps}[FPS]{farthest point sampling}
\acro{gan}[GAN]{generative adversarial network}
\acro{gcn}[GCN]{graph convolutional Network}
\acro{gnn}[GNN]{graph neural network}
\acro{gbot}[GBOT]{graph-based object tracking}
\acro{hmi}[HMI]{Human-Machine-Interaction}
\acro{hmd}[HMD]{head-mounted display}
\acro{dof}[DoF]{degrees of freedom}
\acro{mr}[MR]{mixed reality}
\acro{iot}[IoT]{internet of things}
\acro{icp}[ICP]{iterative closest point}
\acro{icg}[ICG]{iterative correspondence geometry}
\acro{llff}[LLFF]{Local Light Field Fusion}
\acro{bleff}[BLEFF]{Blender Forward Facing}

\acro{lpips}[LPIPS]{learned perceptual image patch similarity}
\acro{nerf}[NeRF]{neural radiance fields}
\acro{nvs}[NVS]{novel view synthesis}
\acro{nfov}[NFOV]{narrow field-of-view}
\acro{mlp}[MLP]{multilayer perceptron}
\acro{mrs}[MRS]{Mixed Region Sampling}

\acro{or}[OR]{operating room}
\acro{pbr}[PBR]{physically based rendering}
\acro{psnr}[PSNR]{peak signal-to-noise ratio}
\acro{pnp}[PnP]{perspective-n-point}
%
\acro{sus}[SUS]{system usability scale}
\acro{ssim}[SSIM]{similarity index measure}
\acro{sfm}[SfM]{structure from motion}
\acro{slam}[SLAM]{simultaneous localization and mapping}

\acro{tp}[TP]{True Positive}
\acro{tn}[TN]{True Negative}
\acro{thor}[thor]{The House Of inteRactions}
\acro{ueq}[UEQ]{User Experience Questionnaire}
\acro{vr}[VR]{virtual reality}
\acro{who}[WHO]{World Health Organization}
\acro{ycb}[YCB]{Yale-CMU-Berkeley}
\acro{yolo}[YOLO]{you only look once}

\end{acronym}


\firstsection{Introduction}

\maketitle

\begin{tikzpicture}[remember picture,overlay]
\node[anchor=south,yshift=10pt] at (current page.south) {\fbox{\parbox{\dimexpr\textwidth-\fboxsep-\fboxrule\relax}{\footnotesize \textcopyright 2024 IEEE. Personal use of this material is permitted. Permission from IEEE must be obtained for all other uses, in any current or future media, including reprinting/republishing this material for advertising or promotional purposes, creating new collective works, for resale or redistribution to servers or lists, or reuse of any copyrighted component of this work in other works. DOI: }}};
\end{tikzpicture}%

Providing error-free assembly of complex assembly groups is highly relevant in manufacturing, maintenance, or medical scenarios~\cite{cramer2024requirement, zauner_authoring_2003,kleinbeck_artfm_2022,murray_equipment_2024}. Object assembly in these cases can be challenging due to the time pressure, the need for high accuracy, and the requirement for prior knowledge about the individual assembly parts.  Moreover, assembly errors in manufacturing can lead to broken parts, extended manufacturing times or in the medical context this can affect the surgery time. Support during assembly tasks can reduce physiological and psychological loads in such scenarios~\cite{tang2003comparative,henderson_2011}. To support these scenarios, \acf{ar} guidance can provide valuable dynamic visualization during the assembly process.  Enabling dynamic visualization for assembly processes requires accurate real-time tracking of the objects and their current state in the assembly. While markers proved high accuracy for tracking in \ac{ar} they are sometimes not suitable for every task in manufacturing or medical scenarios~\cite{cramer2024requirement}. 
An alternative approach to markers in \ac{ar}, is the utilization of 6D pose estimation of assembly objects~\cite{li_gbot_2024}. Deep learning-driven 6D pose estimation determines both the position and orientation of objects in three-dimensional space, enabling markerless tracking. 

To provide in-situ guidance, knowledge of the current assembly state is required along with each objects position in space. Retrieving this knowledge is especially challenging in dynamic assembly scenarios~\cite{zauner_authoring_2003,su_deep_2019}. Furthermore, existing 6D pose estimation approaches~\cite{wang_gdr-net_2021,park_pix2pose_2019} are often limited to static scenes~\cite{jung_housecat6d_2023,wang_phocal_2022} as the benchmark on which they are evaluated are static. Alternatively, object tracking is more often applied in dynamic scenarios with moving objects~\cite{stoiber_iterative_2022,stoiber_multi-body_2022,li_gbot_2024}, but occlusion can be challenging for pure tracking-based approaches~\cite{li_gbot_2024}. 
\ac{ar} assembly guidance combined with 6D pose estimation and state-detection can enable in-situ guidance, and error detection during the assembly to reduce risks and assembly time. Marker-based tracking, however, limits real-world applicability. To enable marker-less 6D pose estimation and state detection, using a deep learning-based approach can be beneficial. 

In summary, existing deep learning-based multi-object approaches a) limit their object and state detection to  2D~\cite{liu_tga_2020,stanescu_state-aware_2023,zhou_fine-grained_2020}, b) limit their evaluation to one or two objects~\cite{liu_tga_2020,murray_equipment_2024}, or c) only provide the 6D object pose information without predicting the current assembly state~\cite{li_gbot_2024,stoiber_fusing_2023,stoiber_iterative_2022}. 

To address these limitations, we present \acf{asdf}, a deep learning-based system for assembly state detection and 6D pose estimation. We build upon the real-time capable \ac{yolo} architecture, fusing 6D pose estimation and assembly state detection for more precise object poses. Additionally, we propose a fully synthetic dataset for training and evaluation, with an additional real-world test scene. Our \ac{asdf} dataset consists of online available 3D printing parts with 6D pose estimation ground truth and assembly state ground truth. The test dataset contains full assembly sequences with hand occlusion and faulty states for a robust evaluation. We evaluate various aspects, such as network size and pose refinement, on our \ac{asdf} dataset.

Furthermore, we compare our approach's 6D pose estimation performance on the \ac{gbot} dataset~\cite{li_gbot_2024}. This evaluation demonstrates the transferability of our domain-randomized training images to the \ac{gbot} evaluation images and shows that assembly state detection can enhance 6D pose performance.

In summary, we contribute:

\begin{itemize}
\item \ac{asdf}, a late fusion approach enhancing assembly state detection and 6D pose estimation through improved assembly state prediction 
\item Our synthetic \ac{asdf} dataset that includes 6D object poses and assembly states using 3D printable parts for reproducibility\footnote{\href{https://github.com/roth-hex-lab/asdf}{GitHub ASDF}}
\item State-of-the-art results on two datasets with assembly assets, demonstrating the advantage of our approach
\end{itemize}

\section{Related Work}

\subsection{6D Object Pose Estimation and Tracking}

Instance-level 6D pose estimation can be divided in one-stager~\cite{he_ffb6d_2021,xiang_posecnn_2017} and two-stager~\cite{wang_gdr-net_2021} approaches. One-stagers are end-to-end trainable~\cite{he_ffb6d_2021,xiang_posecnn_2017}. They extract features from a segmentation or object detection backbone. These can be regressed directly~\cite{xiang_posecnn_2017} or other output like keypoints can be feed to \ac{pnp}~\cite{tekin_real-time_2018}/Least Squares Fitting~\cite{he_ffb6d_2021}. Amini et al.~\cite{amini_YoloPose_2023} introduce YoloPose which directly regresses keypoints in an image and presents a learnable module to replace \ac{pnp}. Two-stagers, apply a state-of-the-art object detection algorithm, for example Faster R-CNN~\cite{girshick2015fast} and build the 6D pose estimation on top of these predictions. Wang et al.~\cite{wang_gdr-net_2021} utilize geometric feature regression on top of the object detection algorithm. This results in 2D-3D correspondences and Surface Region Attention which is leveraged in Patch-\ac{pnp}. Similarly, 
 Pix2Pose~\cite{park_pix2pose_2019} leverages 2D bounding boxes followed by mask prediction step and bounding box refinement step. The final result is predicted using RANSAC \ac{pnp}~\cite{fischler1981random, lepetit2009ep, zheng2013revisiting}. While one-stagers are often more computationally cheap during inference time, two-stagers can provide more accuracy.

In addition to deep learning-based 6D pose estimation, another crucial field for \ac{ar}-driven instructions and object-based applications is object tracking~\cite{VisionLib}. Object tracking starts with a pose initialization and then assumes the 6D object pose in the following frames. Stoiber et al.~\cite{stoiber_iterative_2022} combine visual, regions and depth information. Their improvement ICG$+$~\cite{stoiber_fusing_2023} additionally considers SIFT and ORB features. For objects consisting of multiple parts, Mb-ICG~\cite{stoiber_multi-body_2022} looked at kinematic structures. However, they track the complete assembled object instead of the individual assembly steps. Object tracking can get lost in highly occluded scenes. Li et al.~\cite{li_gbot_2024} propose a combination of an extension of \ac{yolo}v8 and an improvement of \ac{icg} to re-initialize the tracking if it gets lost. While their pose initialization relies on the \ac{cnn}, they utilize an assembly state-graph for tracking assembled objects. Furthermore, they provide the \ac{gbot} dataset for tracking assembled parts. 

\subsection{Assembly State Detection for Augmented Reality}

While the object pose is on essential aspect for assembly guidance another on is the assembly state itself. 
Zauner et al.~\cite{zauner_authoring_2003} utilized markers and build an assembly graph for \ac{ar}-based assembly instruction. \textit{Duplotrack} \cite{gupta_duplotrack_2012} utilizes point cloud alignment whenever the assembly building blocks change. Radkowski et al.~\cite{radkowski2016object}, utilize object tracking with \ac{icp}-based refinment for \ac{ar}-guided assembly matching point clouds.

In terms of 2D assembly state detection, Kleinbeck et al. \cite{kleinbeck_artfm_2022} combine YOLOv5~\cite{jocher_yolo_2023} and synthetic data for building block assembly guidance. The guidance steps are rendered in a Hololens. Similarly, Stanescu et al.~\cite{stanescu_state-aware_2023} provide \ac{ar} glasses-based guidance. For their state detection they introduce an extra convolution block in \ac{yolo}. Overall, their state detection improved the \ac{yolo}-based object prediction. 

Liu et al.~\cite{liu_tga_2020} integrate a two-fold attention mechanism in a 2D object detection architecture.They test their approach on two objects, an IKEA table assembly and Fender assembly. Zhou et al.~\cite{zhou_fine-grained_2020} address mobile \ac{ar} guidance utilizing regions-of-interest for state identification. The method is two-fold, in a first step the regions-of-interest are extracted, in a second step, the state recognition is trained.

Other approaches considering 3D/6D build upon the relative pose between objects~\cite{manuri_state_2019,murray_equipment_2024,wu_augmented_2016}. Wu et al. \cite{wu_augmented_2016} follow a tree like graph structure to determine the assembly state. Similarly, Murray et al.~\cite{murray_equipment_2024} address 6D pose estimation and assembly prediction for robot bin picking defining an assembly state graph. They omit training object detection as a first stage since they are dealing with one single object in front of the camera. For 6D pose estimation they follow Pix2Pose~\cite{park_pix2pose_2019} and address multi-view input utilizing depth estimation and project pixels to 3D.
Su et al.~\cite{su_deep_2019} utilize a TridentNet as backbone and add a pose prediction head. They predict the pose and assembly state for one object with five assembly states and provide \ac{ar} guidance for utilizing the network.

\subsection{Real-world and Synthetic Datasets}

For 6D pose estimation/object tracking multiple benchmarks exist, for example, one common benchmark is the YCB-V benchmark~\cite{xiang_posecnn_2017}. However, this and other common ones considers single objects without state changes~\cite{chao_dexycb_2021,xiang_posecnn_2017,jung_housecat6d_2023,hodan_bop_2020}.
Moreover, providing 6D pose datasets with complex scenes and switching object states is challenging. To address this challenge, one common approach is the use of synthetic data~\cite{schieber_indoor_2024,murray_equipment_2024,li_gbot_2024}. 

\begin{figure*}[t!]
    \centering
   \includegraphics[width=\textwidth]{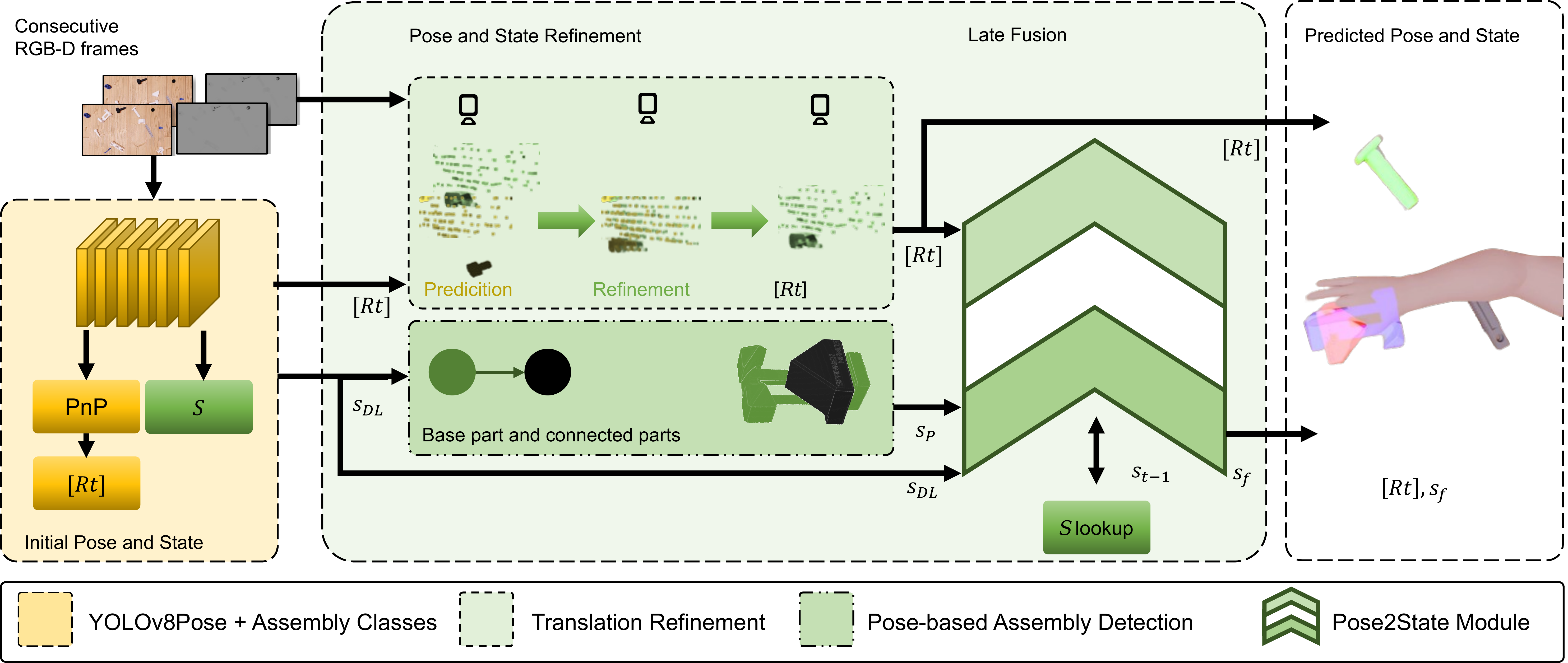}
    \caption{\textbf{Architecture of \ac{asdf}.} We highlight our contribution in green. \ac{asdf} utilizes RGB and depth data. The RGB images are fed into the image backbone and the depth data is used to refine the object poses ([$Rt$]). The image backbone predicts the state ($s$) based on the RGB image. In the Translation Refinement module, the translation offset is calculated. Using the relative pose between the assemblies in the assembly group, we predict a second state assumption in our Pose-based Assembly Detection module. In our final Pose2State module, we weight the individual state predictions to predict the one with the highest probability.}
    \label{fig:arch}
\end{figure*}

\subsubsection{Assembly State Datasets}

One common area for assembly datasets is IKEA furniture~\cite{wang_ikea-manual_2022,su_ikea_2021,liu_tga_2020}. The IKEA-Manual dataset \cite{wang_ikea-manual_2022} provides assembly data including the 3D pose of each part, specifically detailing the rotation of the 3D components. However, it does not explore the translation of individual parts. Addressing this limitation, the IKEA assembly dataset~\cite{su_ikea_2021} fills the gap, although it is not publicly accessible. 

Other works build upon individual objects~\cite{murray_equipment_2024,su_deep_2019,stanescu_state-aware_2023}. However, this makes reproduciblity more challenging. Schoonbeek et al.~\cite{schoonbeek2024industreal} present a hybrid multi-modal dataset focusing on action recognition recorded with a Hololens 2. Su et al.~\cite{su_deep_2019} reconstruct a real-world coffee machine and sample the parts on images to generate a synthetic assembly state dataset. 

For reproducibility, Li et al. \cite{li_gbot_2024} propose the use of 3D printable objects and present a synthetic dataset featuring these objects and their poses. The individual objects have a varying number of assembly states and different sizes.

\subsubsection{Synthetic Data}

The use of synthetic data for 6D position estimation can be useful as obtaining a real markerless dataset is a challenging task~\cite{jung_housecat6d_2023,schieber_indoor_2024,thiel2023automated}. The use of synthetic data poses the challenge of the so-called sim-to-real gap~\cite{schieber_indoor_2024,tobin_domain_2017}. This gap refers to the domain gap between synthetic and real images. To close this gap, various parameters such as light, background, object textures are often randomized and distracting objects are added~\cite{schieber_indoor_2024,tremblay_training_2018,alghonaim_benchmarking_2021}. Tremblay et al.~\cite{tremblay_training_2018} investigated this by combining realistic and randomized images and adding  distractor objects. This increased the performance of object recognition. Alghonhaim et al.~\cite{alghonaim_benchmarking_2021} tested domain randomization considering background, textures and distractors. They also proved that distractors are beneficial for the generalization of a \ac{cnn}.

The constant progress in 6D pose estimation is promising for single objects~\cite{wang_gdr-net_2021,tekin_real-time_2018,peng_pvnet_2019,park_pix2pose_2019,he_ffb6d_2021,zaccaria_self-supervised_2023}, the combination of 6D pose estimation and assembly state detection is even more promising for \ac{ar} guidance approaches~\cite{li_gbot_2024,murray_equipment_2024}. However, for this tasks the used objects hugely vary from building blocks~\cite{tang2003comparative,kleinbeck_artfm_2022,stanescu_state-aware_2023} up to engines with over a hundred assembly states~\cite{murray_equipment_2024}. Existing approaches are often limited to 2D~\cite{stanescu_state-aware_2023}, omit training an object detection model in the first stage~\cite{murray_equipment_2024}, require additional state recognition in addition to their robust object tracking~\cite{li_gbot_2024} or can not handle state switches~\cite{stoiber_multi-body_2022}. To address these limitations we propose \ac{asdf} an end-to-end approach for assembly state detection and 6D pose estimation. Moreover, existing datasets for this task are again either for 2D object detection, 6D pose estimation~\cite{li_gbot_2024}, limited to action recognition~\cite{schoonbeek_industreal_2024} or not publicly available. To enhance comparison in this area we propose our \ac{asdf} dataset containing 6D object pose ground and assembly state truth data using 3D printable object's for comparability and reproducibility.


\section{Method}

\subsection{Assembly State Detection Utilizing Late Fusion}

\ac{asdf} estimates both the 6D pose and the assembly state, which we will explain separately. The individual outputs are fused in a late fusion step within our Pose2State module.

\subsubsection{6D Pose Estimation}

\ac{asdf} uses RGB-D images. The backbone processes the pure RGB image. The Translation Refinement module utilizes the depth image and resulting point cloud to refine the pose, see \autoref{fig:arch}.

For 6D pose estimation, we leverage \ac{yolo}v8Pose~\cite{li_gbot_2024}, an extension of \ac{yolo}v8 utilizing RANSAC \ac{pnp}. To establish keypoints in 3D space, we utilize the 2D keypoints predicted by the backbone. Using \ac{fps}~\cite{li_gbot_2024,peng_pvnet_2019}, we distribute the points as widely apart as possible. The selection of a final number of keypoints (17) creates a balance between computing costs and performance, as more keypoints increase the computing costs.

\paragraph{Assembly Pose Translation Refinement}

Our assembly pose refinement step enhances the estimated translation by identifying the boundary values of each assembly in every assembly group, see \autoref{fig:arch}. To accomplish this, we determine the necessary movement along one coordinate axis (the z-axis) and proportionally adjust the other two axes. To calculate this movement perpendicular to the camera plane, we transform the 3D surface points $P_{3D}$ of the component using the transformation matrix obtained from the initial pose estimation by our \ac{cnn} ($T_{\text{\small{DL}}}$).

The transformed points $P'{3D}$ of the 3D surface points $P{3D}$ using $T_{\small{DL}}$ can be expressed as follows:

\begin{equation}
    P'_{3D} =  \cdot  P_{3D} \cdot T_{\text{\small{DL}}}  
\end{equation}

Next, we back-project these points onto the 2D image plane ($P_{2D}$) and filter them using the predicted bounding box, resulting in visible points ($P'_{2D}$). Subsequently, through a second filtering process, we reduce the number of keypoints to those closest to the camera plane. Based on the 2D points, we approximate the depth information using the depth input. This results in approximated 3D points.

Finally, based on the selected points in $P'_{2D}$ and the approximated depth, we determine the necessary movement along one coordinate axis (i.e., $z$) and proportionally adjust the other two axes (i.e., $x$ and $y$) to refine the translation part. By utilizing the approximated 3D points, we calculate the difference from the corresponding 3D surface points. This allows us to estimate the shift for all points using:

\begin{equation}
    E = \sum_{i=1}^{n} W(d_i) \cdot d_i 
\end{equation}

\( d_i \) denotes the estimated shift distance for point \( i \) and \( n \) represents the total number of points. \( E \) is the overall estimate of the required shift for all points.

To estimate the shift distance for each point, we use \( d_i \). Then, to obtain an overall estimate, we combine these individual estimations using a weighting function \( W(d_i) \). This function assigns higher weights to points with smaller differences, thereby preventing occluded objects from influencing false calculations during the refinement step.

The resulting translation along the axis perpendicular to the image plane can now be utilized to determine the vector along the camera view axis, representing the final translation.

\subsubsection{Assembly State Detection}

The final assembly state detection combines three key components: deep learning-based state detection, relative pose-based state detection, and consideration of the previous state using a weighted late fusion.

\paragraph{Deep learning-based Assembly State Detection}

The relationship between assembly state and 6D pose estimation is interdependent. Each pose of an assembly part within a group contributes valuable information to the overall assembly state, and conversely, the current assembly state informs the relative poses of the objects to each other.

To integrate state detection in \ac{asdf}s backbone, we represent each state as a distinct class. The assembled parts with a new assembly state are regarded as a new class. 

\paragraph{Pose-based Assembly State Detection}

The relative position between objects provides valuable information about the current assembly state of each assembly. These relative poses are known because they are crucial for the creation of training and evaluation data.

To determine the assembly state based on the relative pose of an assembly, we select the base part of each assembly. Followed by this, we can interpret the relative poses of the entire assembly.

\subsubsection{Pose2State Module}

Given the inherent co-dependency between pose and state, we harness this relationship in our Pose2State module in a late fusion manner. This module is designed to forecast the state probability $(SP)$ of the assembly state $s_x$ at time $t$. Our Pose2State module seamlessly integrates two key components: the deep learning-based prediction $(SP_{\small{DL}})$ and the prediction derived from the pose-based assembly $(SP_{\small{P}})$. By combining these predictions, our module effectively determines the final assembly state, leveraging both deep learning and the pose-based assembly state.

We first combine the deep learning-based state prediction $(SP_{\small{DL}})$ and the pose-based state prediction $(SP_{\small{P}})$:

\begin{equation}
    SP_{\small{DL+P}}(s_x)_{t} = w_{DL} \cdot P_{\small{DL}}(s_x)_t + w_P \cdot P_{\small{P}}(s_x)_t
\end{equation}

For the final assembly state $SP_{f}$ we consider the time component and leverage the previously stored assembly state (${(s_x)}_{t-1}$). $P_{f}(s_{x\pm1})_{t-1}$ 
represents the average normalised probability of the two neighbouring states $s_{x\pm1}$ from the previous time step $(t-1)$.

\begin{equation}
    SP_{\small{f+f_{t-1}}}(s_x)_{t} = w_f \cdot P_{\small{f}}(s_x)_{t-1} + w_{f-1} \cdot P_{\text{\small{f}}}(s_{x\pm1})_{t-1}
\end{equation}

This results in the probability ($P$)  for the final assembly state $SP_{\small{f+f_{t-1}}}(s_x)_{t}$ at time $t$. We fuse the state probability ($SP$) our deep learning-based and relative-pose based estimation ($SP_{\small{f+f_{t-1}}}(s_x)_{t}$) and weight them:
    
\begin{equation}
 P_f(s_x)_t = \frac{SP_{\text{\small{DL+P}}}(s_x)_{t} + SP_{\small{f+f_{t-1}}}(s_x)_{t}}{w_{DL} + w_{P} + w_{f} + w_{f-1}}
 \label{eq:state_calculation}
\end{equation}

\subsection{Assembly State Dataset}

\begin{figure}[t!]
    \centering
    \includegraphics[width=\columnwidth]{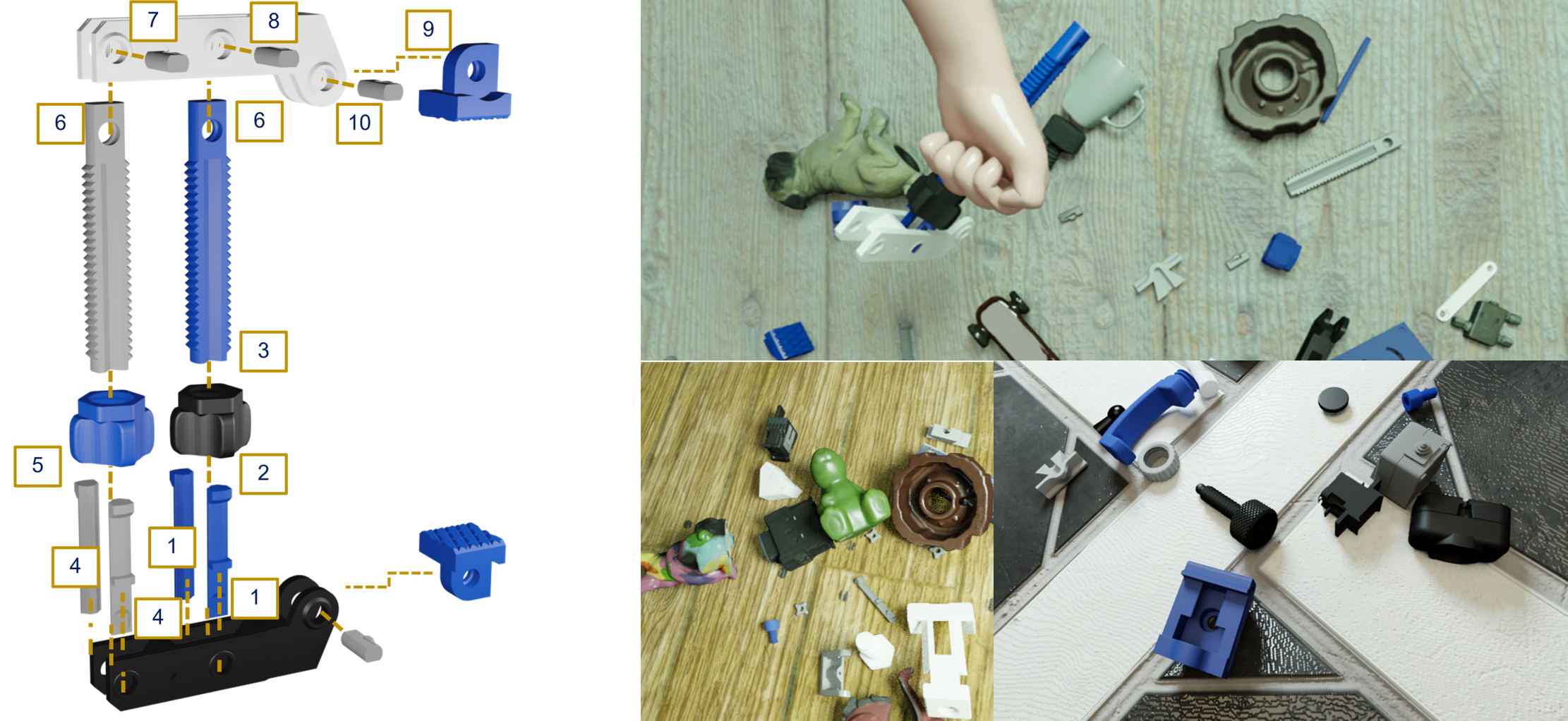}
    \caption{Assembly state complexity of the \textbf{\ac{asdf} dataset  (left) and training images of the \ac{asdf} dataset (right).} For training we use assembled and unassembled data (right) and additionally provide hand occlusion (top-right), varying background and light conditions as well as distracting objects. We include the state information in our  ground truth labels. An example of the state complexity can be seen in the left figure. \label{fig:syn-training}}
    
\end{figure}

To address assembly states and 6D pose estimation, we created a synthetic dataset\footnote{\href{https://github.com/roth-hex-lab/asdf}{GitHub ASDF}} with 3D-printed parts to facilitate testing under real conditions. We used the 3D printable parts from the \ac{gbot} dataset~\cite{li_gbot_2024} as a foundation. In addition to the existing reproducible setup~\cite{li_gbot_2024}, we introduced assembly state information, wrong assembly states and incorporated hand occlusion during training to enhance the model's robustness. Our dataset adheres to the camera specifications of the Azure Kinect DK, with a resolution set to $1280 \times 720$ pixels. The camera is positioned in a top-down view to simulate a capturing setup commonly found in medical or industrial scenarios.

\subsubsection{Synthetic Dataset}
     \begin{table}[t]
         \centering
         \caption{\textbf{Number of states and synthetic test images for each assembly part.} The assembly groups of our \ac{asdf} dataset (left column), the number of states (center column), and number of images in the test set.}
         \begin{tabular}{l|ccc} 
         \toprule
         \textbf{Assembly} & \textbf{No. States} & \textbf{Number of images}  \\ 
         \midrule
         NanoVise  &  8  & 191  \\
         {ScrewClamp}  & 10 & 231  \\
         {GearedCaliper}  & 5  & 111  \\
         {CornerClamp} &  3  & 66\\
         \bottomrule
         \end{tabular}
         \label{tab:no_eval_images}
     \end{table}

\paragraph{Training and Validation Dataset}

\begin{figure}[t!]
    \centering
    \includegraphics[width=\columnwidth]{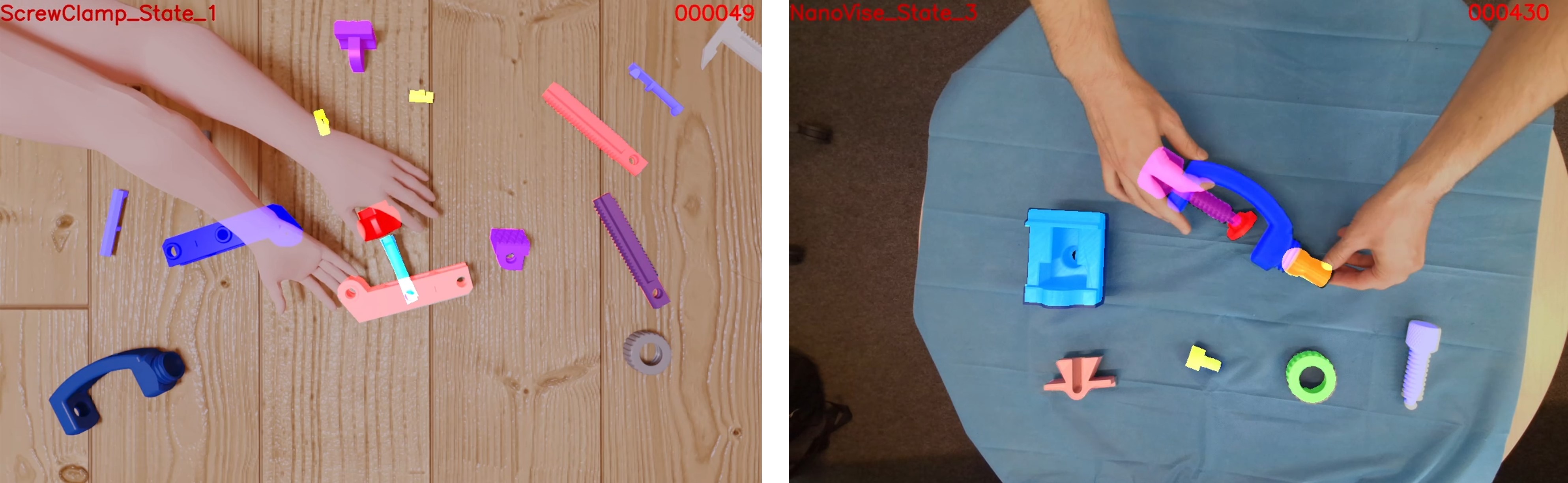}
    \caption{\textbf{Example images with highlighted ground truth of our \ac{asdf} test set.} Synthetic image (left) and real-world image (right). The ground truth of each currently evaluated assembly group is visualized with colorful overlays. }
    \label{fig:real-syn}
\end{figure}

Our synthetic dataset includes 6D pose estimation and assembly state detection ground truth data with 20k images per assembly, using an 80:20 training split. For testing each test scene contains a various number of images with several levels of domain randomization, see \autoref{tab:no_eval_images}. Our dataset is generated using BlenderProc~\cite{denninger_blenderproc_2020}. Domain randomization is crucial in synthetic data generation \cite{schieber_indoor_2024,alghonaim_benchmarking_2021}. Thus, we introduce distracting objects \cite{hodan_t-less_2017,kaskman_homebreweddb_2019} and simulate hand occlusion to reflect real-world scenarios, as shown in \autoref{fig:real-syn}. In addition to, to distracting objects and hand occlusion, we add randomized noise, different lighting variations, and varying background materials~\cite{alghonaim_benchmarking_2021,schieber_indoor_2024}. 

\paragraph{Test Dataset}
For each assembly we generate a continuous sequence of assembly images resulting in a full assembly video. The test scenes contain one full assembly sequence including error states to test the robustness of the assembly state detection. We maintain the presence of distracting objects and incorporate hands at realistic assembly positions to mimic occlusion and test robustness. The objects are randomly distributed within the camera's field of view. Each frame contains assembly state and 6D pose ground truth data.

A key aspect of our test set is the inclusion of incorrectly assembled states. In a real-world assembly scenario mounting together parts with missing pieces in between can happen especially if the parts are not necessarily connecting parts. Therefore, we added incorrect assembly states to test robustness. Since each object contains various assembly states, the number of evaluation images per assembly object varies, as illustrated in \autoref{tab:no_eval_images}. This variability ensures comprehensive testing of the model's ability to detect and handle different assembly scenarios.
 
\section{Evaluation}

To benchmark the accuracy of the assembly state detection and 6D pose estimation, we evaluate our approach on our \ac{asdf} dataset and on the \ac{gbot}~\cite{li_gbot_2024} datasets. Other comparisons with existing assembly state detection approaches~\cite{stanescu_state-aware_2023,liu_tga_2020} in 2D are not possible, since both pose and state are relevant for the final prediction. Other works focus purely on state changes~\cite{schoonbeek_industreal_2024}, purely on 6D pose performance~\cite{li_gbot_2024} or build upon non publicly available~\cite{su_deep_2019,murray_equipment_2024} assets or data.

\paragraph{\ac{gbot} Dataset:} The \ac{gbot} dataset~\cite{li_gbot_2024} is a exclusively a synthetic dataset designed for 6D object pose/tracking tasks. It comprises 3D printable assembly parts and contains five evaluation sequences, each presenting different levels of difficulty for object tracking (normal (N), dynamic (D), hand occlusion (H), and blur (B)). Utilizing the \ac{gbot} dataset, we conduct a comprehensive comparison of our approach against state-of-the-art methods in 6D pose estimation performance, particularly focusing on assemblies with dynamically changing states.

\subsection{Implementation and Specifications}

We trained our network for 300 epochs using early stopping. The training for all comparisons is executed on one machine with an Intel Core i9-10980XE CPU, 128 GB RAM and one NVIDIA GeForce RTX 3090 graphics card. 

As loss term, we follow \ac{yolo}v8 combining the class-wise loss ($L_{cls}$), bounding box loss ($L_{box}$), the task-specific loss ($L_{tsk}$) all weighted by $\lambda$. 

$L_{tsk}$ is comprised of the pose loss (${L}_{pose}$) which is defined as ${L}_{pose} =|| \mathbf{K}_{pred} - \mathbf{K}_{gt} ||_2$.
This is the L2 loss using the predicted keypoints (${L}_{pose}$) from the ground truth keypoints ($\mathbf{K}_{gt}$) and the Cross-Entropy (CE) Loss using the keypoints.

\begin{equation}
L_{total} = \sum_{i,j,k}(\lambda_{cls}L_{cls} + \lambda_{box}L_{box} + \lambda_{tsk}L_{tsk})
\end{equation}

\subsection{Metrics}

We evaluate our approach based on two main aspects: 6D pose estimation and state detection. For pose prediction, we report the absolute translation ($e_t$)~\cite{hodan_evaluation_2016}, rotation error $e_r$~\cite{hodan_evaluation_2016} and \ac{add}(S)~\cite{hinterstoisser_model_2013,li_gbot_2024}. \ac{add}(S) describes the average distance error for asymmetric (\ac{add}) and symmetric (\ac{add}(S)).

For state detection accuracy, we calculate the F1 score. The F1 score is composed of the calculation of precision and recall for each state ($s$) and each assembly part ($c$).

Since our goal is to utilize our state and pose recognition, we also evaluate the runtime. For the runtime, we calculate the average value over all assembly objects.

\subsection{ASDF Results} 

We evaluate performance trade-off using different networks sizes, pose accuracy considering various refinement steps and benchmark our final approach.

\subsubsection{Performance Trade-off}

Since \ac{asdf} leverages object pose information to refine the state, our initial focus lies on evaluating the pose performance. As depicted in \autoref{tab:runtime_asdf}, it is evident that the fastest network performance can be attained using the smallest network configuration, which is a rational expectation. As overall network architecture we build upon \ac{yolo}v8Pose~\cite{li_gbot_2024}. However, the underlying image backbone can vary in terms of network size. As shown in \autoref{tab:runtime_asdf}, the individual size \texttt{n}, \texttt{m}, \texttt{l} and \texttt{xl-p6} have all their individual advantages and disadvantages for the pose performance. \texttt{m} and \texttt{l} exhibit similar performance, whereas the \texttt{xl-p6} architecture, serving as our underlying backbone, demonstrates the most favorable results in terms of rotation and translation errors.

\begin{table}[t!]
\caption{\textbf{Ablation study on the impact of pose refinement steps and runtime on our \ac{asdf} dataset.} We compare our keypoint-based \ac{asdf} with YoloV8Pose + State + ICP (ICP-based refinement), and use an additional YOLO-based segmentation network for refinement (+ Seg). We provide a comparison using the smallest backbone (n) and the final backbone of \ac{asdf} (xl-p6). The best performing approach per category is \textbf{bold}. \label{tab:pose_performance}}
\resizebox{\columnwidth}{!}{
\begin{tabular}{ll|ccc} \toprule
\textit{Method}& \textit{Backbone}  & Runtime {[}ms{]}  $\downarrow$   & $e_{trans} $ $\downarrow$  {[}mm{]}  $\downarrow$ & erot {[}°{]}   $\downarrow$  \\ \midrule
\multicolumn{2}{l}{\textit{No refinement}}  & \\ \midrule
~~\ac{yolo}v8Pose + state & \texttt{n}    & \textbf{24.83}& 29.13 & 18.76  \\
~~\ac{yolo}v8Pose + state & \textbf{xl-p6}& 50.83  & {19.11}  & \textbf{11.14} \\ \midrule
\textit{ICP refinment}       &&  & \\  \midrule
~~\ac{yolo}v8Pose + state + ICP & \texttt{n}    & 151.77 & 12.27& 18.96  \\
~~\ac{yolo}v8Pose + state +ICP & \texttt{xl-p6} & 145.76 & {9.24}   & {14.35} \\  \midrule
\multicolumn{2}{l}{\textit{Segmentation refinement}} &  & \\  \midrule
~~\ac{asdf} + Seg & \texttt{n}     & 64.46     & 8.52 & 18.76  \\
~~\ac{asdf} + Seg & \texttt{xl-p6} & 85.03  & 6.53 & \textbf{11.14} \\  \midrule
\multicolumn{2}{l}{\textit{Ours}} &  & \\  \midrule
~~\ac{asdf} & \texttt{n}   & {32.45} & 8.13 & 18.76  \\
~~\textbf{\ac{asdf}} & \texttt{xl-p6} &  {55.70}   & \textbf{5.98}   & \textbf{11.14} \\ \bottomrule
\end{tabular}}
\end{table}

\begin{table}[t!]
\centering
\caption{\textbf{Ablation study on the impact of the underlying network size on the pose accuracy.} We compare the individual backbone network sizes of \ac{yolo}v8Pose. As metrics we report runtime in ms and pose translation error in mm and rotation error in degree on the \ac{asdf} dataset. The best performing approach per category is \textbf{bold}. \label{tab:runtime_asdf}}
\begin{tabular}{l|ccc} \toprule
\textit{Backbone}  & Runtime {[}ms{]}    $\downarrow$    & $e_{trans} $ $\downarrow$  {[}mm{]}  $\downarrow$ & erot {[}°{]}   $\downarrow$ \\ \midrule
\texttt{n}    & \textbf{24.83}& 29.13& 18.76  \\
\texttt{m}    & 33.72  & 27.85& 15.21  \\
\texttt{l}    & 35.55  & 23.69& 13.09  \\
\textbf{xl-p6}& 50.83  & \textbf{19.11}  & \textbf{11.14} \\ \bottomrule
\end{tabular}
\end{table}

\begin{figure*}[t!]
    \centering
   \includegraphics[width=1\textwidth]{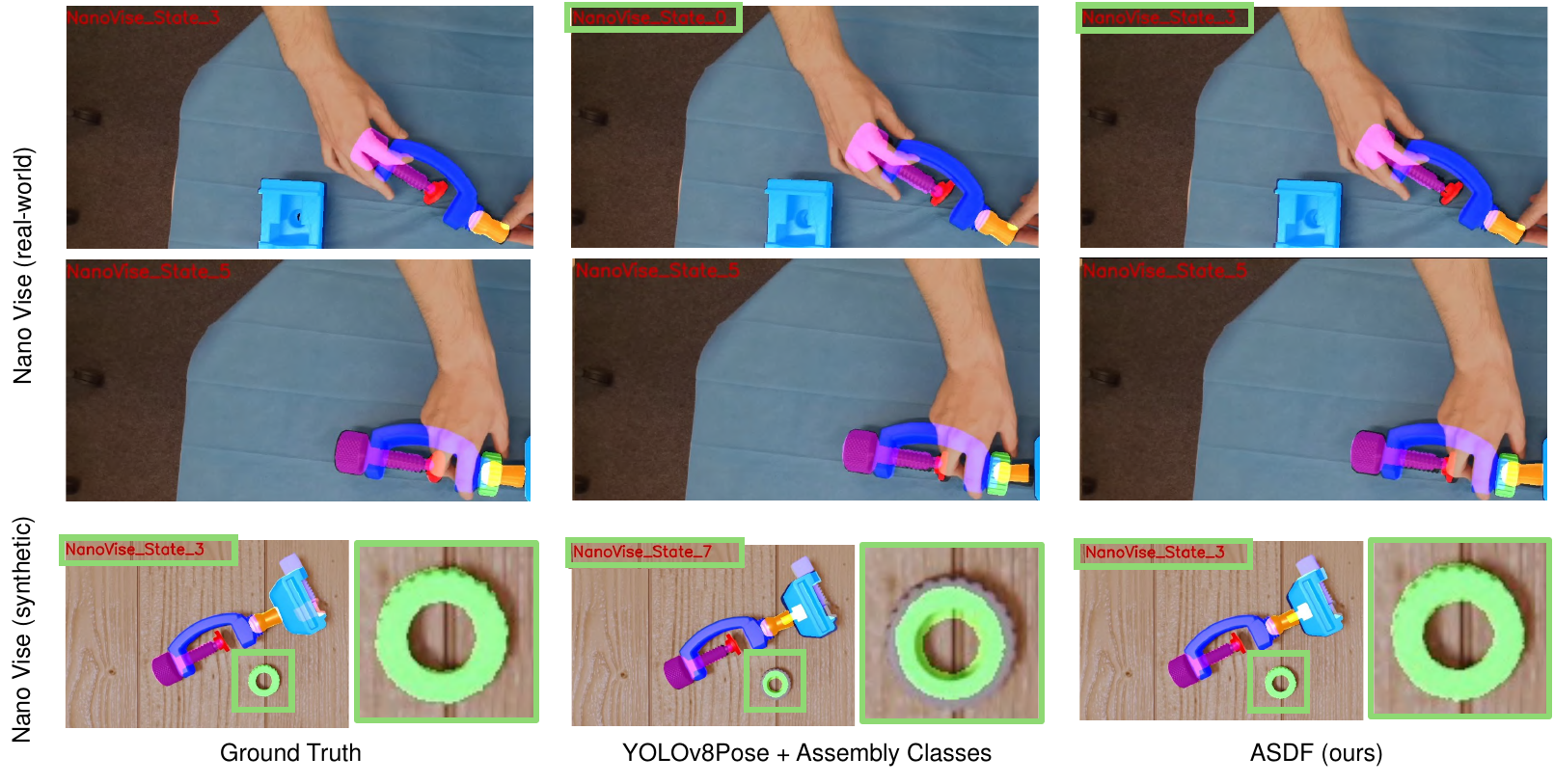}
    \caption{\textbf{Example of the results on the \ac{asdf} test set.} We show the performance of \ac{asdf} compared to \ac{yolo}v8Pose + Assembly Detection on real-world captures (top two lines) and synthetic renderings (bottom line). The ground truth (left), the pure YOLOv8-based pose and state prediction (center) and our prediction using \ac{asdf}  (right). The current predicted state is denoted in every top-left corner and the pose is shown with a colorful overlays.}
    \label{fig:asdf_prediction_state_pose}
\end{figure*}

\begin{figure}[t!]
    \centering
    \includegraphics[width=0.95\columnwidth]{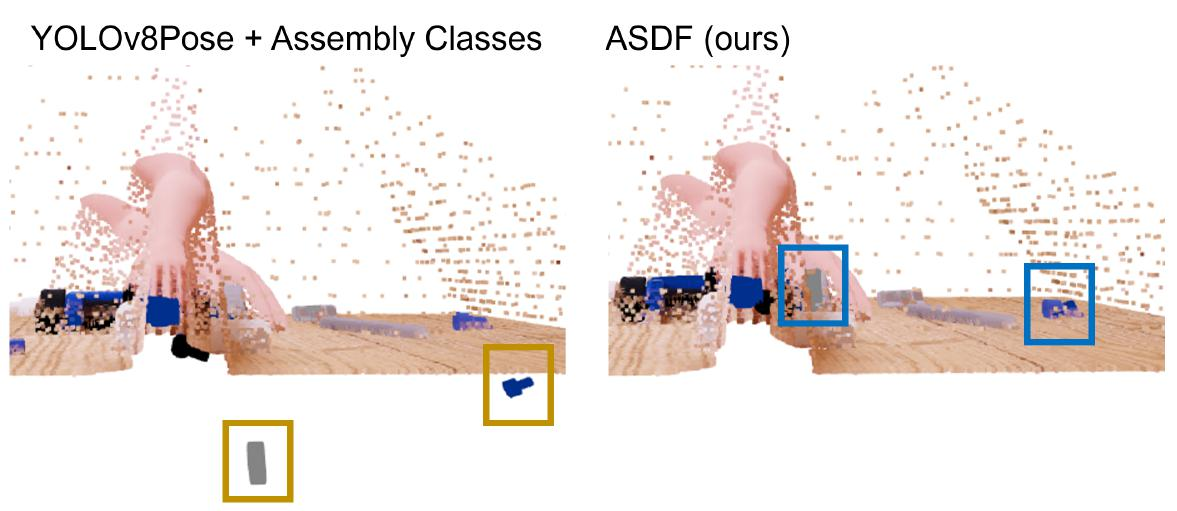}
    \caption{\textbf{Example comparison of the translation offset using \ac{yolo}v8Pose and \ac{asdf}.} \ac{yolo}v8Pose shows an offset in 3D (left, yellow) while the translation refinement of our \ac{asdf} (right, blue) can address this shift.\label{fig:pose-ref}}
    
\end{figure}

\begin{table*}[t!]
    \centering
    \caption{\textbf{Results on the \ac{asdf}~dataset}: The \ac{add}(S)($\uparrow$) is calculated with $10cm$ threshold, the translation error is in milimeters (mm) denoted as $e_{trans} $ [mm], $\downarrow$) and the rotation error denoted as $e_{rot} [^{\circ}]$, $\downarrow$) in degrees. The best results among all methods are labeled in \textbf{bold}. We compare \ac{yolo}v8Pose using deep learning-based state detection with \ac{asdf}. To report the state performance we report the F1 score. \label{tab:all_asdf}}
    \resizebox{0.8\textwidth}{!}{
    \begin{tabular}{l|cccc|cccc} \toprule
    & \multicolumn{4}{c|}{\ac{yolo}v8Pose + Assembly Classes}   & \multicolumn{4}{c}{ASDF}     \\ \midrule
    Assembly & F1  $\uparrow$ & $\acs{add}(S) $  $\uparrow$ & $e_{trans} $ [mm]  $\downarrow$ & $e_{rot} [^{\circ}]$  $\downarrow$ & F1  $\uparrow$ & $\acs{add}(S) $  $\uparrow$ & $e_{trans} $ [mm]  $\downarrow$ & $e_{rot} [^{\circ}]$  $\downarrow$ \\ \midrule
    Corner Clamp    &   87.88	& 89.22   & 15.74   &	\textbf{8.51}  &	\textbf{93.85} & \textbf{95.69} & \textbf{5.39}	& 8.76 \\
    Screw Clamp     &   73.28   & 85.53   &	23.07	& 22.48	  & \textbf{84.35} & \textbf{95.02} & \textbf{9.12} & \textbf{20.68} \\
    Geared Caliper  &   58.77   &	81.11 &	{22.92} & \textbf{16.69} & \textbf{60.91} & \textbf{96.33}	& \textbf{4.65} & {17.03}  \\
    Nano Vise       &   75.45   & 83.82   & 21.88   & \textbf{5.04} & \textbf{78.95} & \textbf{95.92}	& \textbf{7.63} & {8.22} \\ \midrule
    Mean            &   {73.85}	& 84.92   &	20.90   & \textbf{13.18} &	\textbf{79.52} & \textbf{95.74} & \textbf{6.70}	& {13.67}  \\ \bottomrule
    \end{tabular}}
\end{table*}

\begin{table*}[t!]
\centering
    \caption{\textbf{Results on the \ac{gbot}~dataset:} We compare our approach on the \ac{gbot} benchmark with the four conditions normal (N), dynamic (D), hand (H) and blur (B). We compare against their pure deep learning-based approach, the combined approach of deep learning and object tracking (\ac{gbot} + re-init) and pure tracking-based approaches. Rotational errors are only evaluated for unsymmetrical objects. The \ac{add}(S)($\uparrow$) is calculated with $10cm$ threshold, the translation error is in millimeters (mm) ($e_{ave\_{trans}}$ denoted as $e_{trans}$, $\downarrow$) and the rotation error ($e_{ave\_ rot}$ denoted as $e_{rot}$, $\downarrow$) in degrees. The best results among all methods are labeled in bold. In this evaluation, tracking is initialized with ground truth pose only for the first frame and the pose is not reinitialized afterwards. In the last column, the tracking results of tracking re-initialization by pose estimation are shown. \label{tab:gbot}}

    \resizebox{\textwidth}{!}{

    \begin{tabular}{ll|ccc|ccc|ccc|ccc|ccc|ccc|ccc} 
    \toprule
    \multicolumn{2}{c|}{Approach} & \multicolumn{6}{c|}{6D pose estimation}& \multicolumn{12}{c|}{Tracking}& \multicolumn{3}{c}{6D pose estimation + tracking} \\ 
    \midrule
    \multirow{3}{*}{Asset}& \multirow{3}{*}{} & \multicolumn{3}{c|}{} & \multicolumn{3}{c|}{} & \multicolumn{3}{c|}{} & \multicolumn{3}{c|}{} & \multicolumn{3}{c|}{} & \multicolumn{3}{c|}{} & \multicolumn{3}{c}{} \\
    &&  \multicolumn{3}{c|}{\acs{asdf} (ours)} & \multicolumn{3}{c|}{\acs{yolo}v8Pose~\cite{li_gbot_2024}} &\multicolumn{3}{c|}{SRT3D~\cite{stoiber_srt3d_2022}}  & \multicolumn{3}{c|}{ICG~\cite{stoiber_iterative_2022}} & \multicolumn{3}{c|}{ICG+SRT3D~\cite{li_for_2023}}& \multicolumn{3}{c|}{GBOT~\cite{li_gbot_2024}}& \multicolumn{3}{c}{\ac{gbot} + re-init~\cite{li_gbot_2024}} \\   
    & & \multirow{2}{*}{\footnotesize{$\acs{add}(S) $}}  & $e_{trans} $ $\downarrow$ & $e_{rot} $  $\downarrow$  & \multirow{2}{*}{\footnotesize{$\acs{add}(S) $}} & $e_{trans} $  $\downarrow$ & $e_{rot} $  $\downarrow$  & \multirow{2}{*}{\footnotesize{$\acs{add}(S) $}} & $e_{trans} $  $\downarrow$ & $e_{rot} $  $\downarrow$ & \multirow{2}{*}{\footnotesize{$\acs{add}(S) $}} & $e_{trans} $ $\downarrow$ & $e_{rot} $  $\downarrow$  & \multirow{2}{*}{\footnotesize{$\acs{add}(S) $}} & $e_{trans} $ $\downarrow$ & $e_{rot} $  $\downarrow$  & \multirow{2}{*}{\footnotesize{$\acs{add}(S) $}} & $e_{trans} $   $\downarrow$ & $e_{rot}$  $\downarrow$ & \multirow{2}{*}{\footnotesize{$\acs{add}(S) $}} & $e_{trans} $  $\downarrow$ & $e_{rot} $  $\downarrow$   \\
    & & &  [mm]  & $[^{\circ}]$ & &  [mm]  & $[^{\circ}]$ & &  [mm]  & $[^{\circ}]$ &  & [mm]  & $[^{\circ}]$  & & [mm]  & $[^{\circ}]$ & &  [mm]  & $[^{\circ}]$ &  &  [mm]  & $[^{\circ}]$  \\
    \midrule
    & N & \textbf{100.0}& 18 & 3.0 & 91.5 & 93 & 3.8 & 89.8 & 47& 25.8& \textbf{100.0}& 1& 38.7& \textbf{100.0}& 13 & 38.7& \textbf{100.0}& 6 & \textbf{2.1} & 99.5 & 7 & 2.7  \\  
    Corner & D  & \textbf{100.0}& 18 & \textbf{2.7} & 99.0 & 25 & 4.8 & 88.4 & 37& 25.0& \textbf{100.0}& 15& 46.8& \textbf{100.0}& 20& 47.8& \textbf{100.0}& 6& 23.2& \textbf{100.0}& \textbf{5}& 3.5  \\ 
    Clamp & H & 68.6 & 83 & 38.4&  45.4 & 541 & 97.1 & 66.6 & 88 & \textbf{37.1}& 68.4 & 74 & 59.4 & 68.4 & 75 & 58.3& 81.9 & 57 & 90.4& \textbf{90.6} & \textbf{44}& 84.7 \\  
    & B  & \textbf{100.0}& 19 & 3.4 & 97.3 & 40 & 4.3 & 88.9 & 50& 30.4& \textbf{100.0} & \textbf{5}& 2.2 & \textbf{100.0}& 10 & 3.0 & \textbf{100.0}& 6 & \textbf{2.1} & 99.9 & 6& \textbf{2.1}  \\ 
    \midrule
    & N& \textbf{100.0}& 4 & 4.7 & 99.6 & 14 & 10.1 & 90.6 & 18 & 5.4 & \textbf{100.0}& \textbf{2}& 2.6 & \textbf{100.0}& 3& 3.0 & \textbf{100.0}& \textbf{2}& \textbf{2.1} & \textbf{100.0}& 5 & 3.3  \\ 
    Geared& D  & \textbf{100.0}& 3 & 4.1 & 99.9 & 13 & 9.3 & 92.7 & 14& 9.9 & \textbf{100.0}& \textbf{2}& 2.5 & \textbf{100.0}& 0.4& 3.6 & \textbf{100.0}& \textbf{0.2}& \textbf{2.3} & \textbf{100.0}& 0.5& 3.6  \\  
    Caliper& H  & \textbf{100.0}& 8 & 6.9 & 99.2 & 24 & 12.5 & 96.5 & 9& \textbf{4.4} & 85.4 & 30 & 30.0& 85.5 & 31 & 30.0 & 85.4 & 30& 30.0& 99.6 & 8 & 7.0  \\ 
    & B  & \textbf{100.0}& 4 & 4.9 & 99.6 & 14 & 9.9  & 98.9 & 9& 8.7 & \textbf{100.0}& \textbf{2}& 2.5 & \textbf{100.0}& 3& 2.8 & \textbf{100.0}& \textbf{2}& \textbf{2.2} & \textbf{100.0}& 5& 3.6  \\ 
    \midrule
    & N & \textbf{100.0}& 14 & 4.8 & 71.1 & 88 & 4.8  & 74.1 & 66& 13.8& 89.8 & 21& 16.7& 89.4 & 23& 16.5& 99.8 & \textbf{6}& 7.2 & 93.8 & 24& \textbf{3.6}  
    \\ 
    Nano& D & 90.0 & 16& 5.0 & 72.6 & 76 & 4.6 & 63.4 & 103 & 15.6& 87.3 & 39 & 15.3 & 87.3 & 41 & 15.8 & 96.0 & 24& 20.0 & 92.9 & 25& \textbf{3.7}  \\  
    Vise& H  & 98.6 & 19& \textbf{5.3} & 65.9 & 110 & \textbf{4.7} & 61.4 & 136  & 15.3& 76.5 & 60 & 18.3& 75.8 & 61 & 18.1& 72.9 & 87& 14.5& 87.8 & 31& 7.3  \\  
    & B & \textbf{100.0}& 15& 5.4  & 70.9 & 99 & 5.2& 61.9 & 116  & 15.2 & 91.6 & 19& 11.4& 91.5 & 21& 11.1& 95.7 & \textbf{7}& 25.1& 92.7 & 30& \textbf{4.7}  \\ 
    \midrule
    & N& 93.5 & 11& 8.6 & 73.5 & 77 & 17.6 & 86.5 & 47& 8.9 & 96.0 & 12& 1.1 & 95.9 & 14& 2.3 & \textbf{98.8} & \textbf{4}& \textbf{0.9} & 83.7 & 30& 4.7  \\ 
    Screw & D & 94.5 & 11& 9.4 & 82.0 & 67 & 18.4 & 86.4 & 56& 27.0& 95.9 & 17& 2.1 & 95.9 & 22& 3.4 & \textbf{98.8} & \textbf{6}& \textbf{1.7} & 91.6 & 21& 6.1  \\  
    Clamp& H & \textbf{92.1} & \textbf{18}& \textbf{13.5} & 67.2 & 142 & 39.8 & 60.1 & 143  & 37.7& 73.4 & 69& 53.4& 73.1 & 71& 53.4& 68.7 & 79& 61.9& 83.9 & 49& 27.2 \\ 
    & B  & 96.1 & \textbf{9} & 8.7 & 85.6 & 65 & 18.6 & 86.5 & 56 & 34.6 & 95.7 & 30 & 6.6 & 95.7 & 32& 7.7 & \textbf{98.6} & 12& \textbf{1.1} & 91.1 & 30 & 4.8  \\ 
    \midrule
    Mean&  &\textbf{95.8} & \textbf{17} & \textbf{8.1} & 82.52 &	93 &	16.59
 & 80.8 & 62& 19.7& 91.3 & 25& 19.4& 91.2 & 28& 19.7& 93.5 & 21& 17.9& 94.2 & 20 & 10.8 \\
    \bottomrule
    \end{tabular}
    }
\end{table*}

\subsubsection{Pose Refinement}

The backbone \texttt{n} shows the best runtime and the network size \texttt{xl-p6} the best performance, see \autoref{tab:runtime_asdf}. Therefore, we further compared these two backbones considering different pose refinement methods. \ac{asdf} builds upon the use of keypoints for the 6D pose estimation. However, we considered other refinement methods as well to lift the pose performance. In \autoref{tab:pose_performance} we compare the pure YOLOv8pose combined with the state prediction with the commonly used \ac{icp}-based refinement (YOLOv8pose + state + \ac{icp}). Additionally, since semantic segmentation provides clearer object boundaries, we compared our keypoint-based approach using a refinement step with a second YOLOv8~\cite{jocher_yolo_2023} semantic segmentation network and our ASDF with the backbones \texttt{n} and  \texttt{xl-p6}.

The improvement of our translation refinement compared to the pure network-based output can be seen in \autoref{fig:pose-ref} and in \autoref{tab:runtime_asdf}. As shown in \autoref{tab:pose_performance} neither the additional segmentation network nor \ac{icp}-based pose refinement can outperform \ac{asdf}.

\subsubsection{Assembly State Detection and 6D Pose Estimation}

As shown in \autoref{tab:runtime_asdf} and \autoref{tab:pose_performance}, \ac{asdf} shows promising results in terms of 6D pose estimation. Comparing the assembly state detection using the F1 score, see \autoref{tab:all_asdf}, shows that \ac{asdf} can predict the assembly state better (79.52) compared to a pure deep learning-based approach  (73.85). Moreover, as denoted in \autoref{tab:all_asdf}, the 6D pose performance shows an improved performance by outperforming the deep learning-based approach in \ac{add}(S) (95.74 (ours) compare to 84.92). Moreover, for the translation error we can reduce the average error by 14.2 mm.

For qualitative analysis we selected images from our test set, see \autoref{fig:asdf_prediction_state_pose}.  It becomes apparent that, the state detection of \ac{asdf} is more precise compared to the deep learning-based approach. As shown in \autoref{fig:asdf_prediction_state_pose} on different assemblies in different states the predicted poses are more accurate as well for assembled (Screw Clamp) and unassembled (Nano Vise, Caliper) parts.

Moreover, we 3D printed the real-world assembly and manually annotated one sequence. As shown in \autoref{fig:asdf_prediction_state_pose}, our approach allows the detect the correct state during a real-world state transition.

\begin{figure}[t!]
        \centering
        \includegraphics[width=\columnwidth]{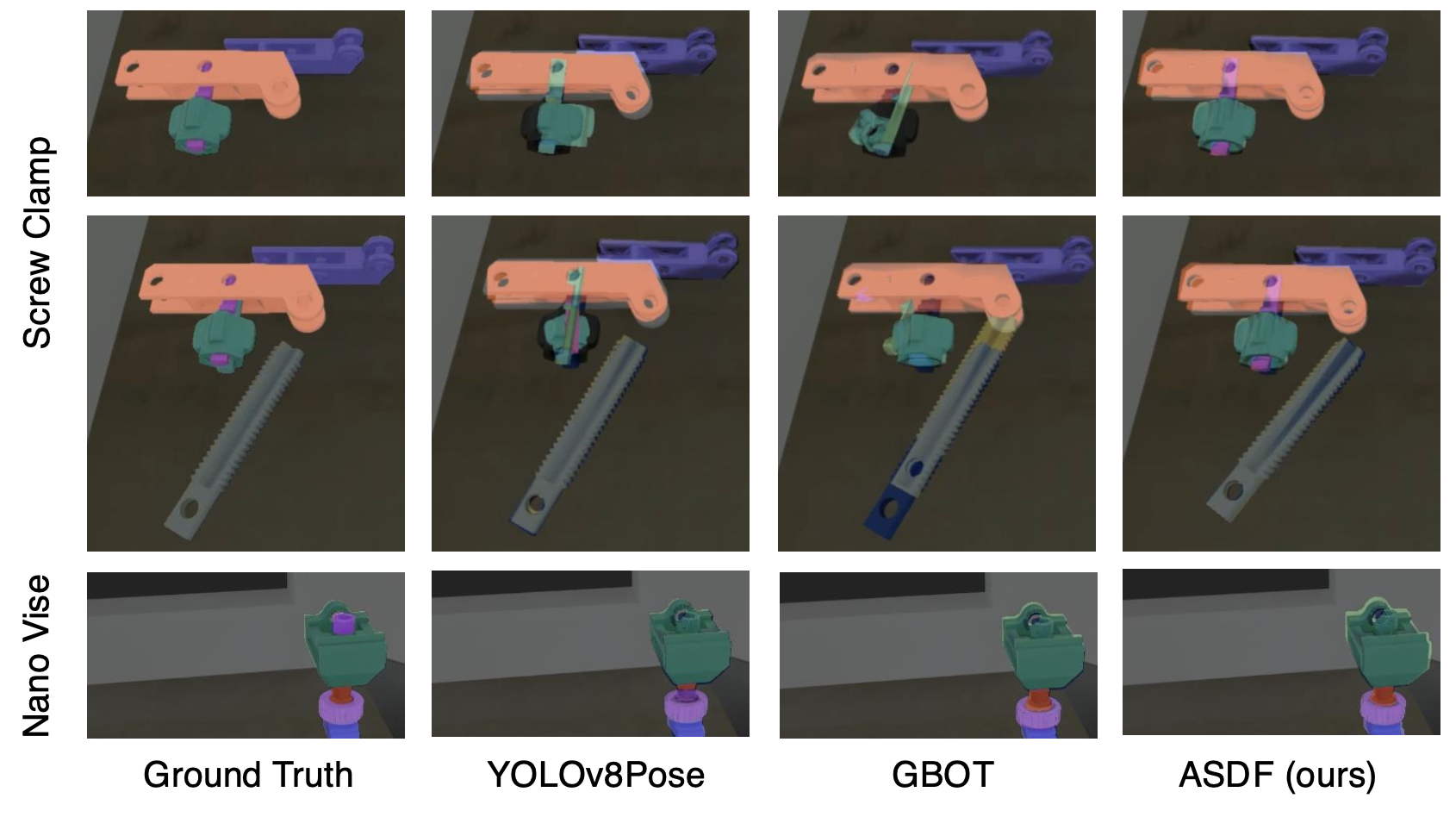}
        
        \caption{\textbf{Qualitative comparison of \ac{yolo}v8Pose~\cite{li_gbot_2024}, \ac{gbot}~\cite{li_gbot_2024} and \ac{asdf} (ours) on the \ac{gbot} dataset}: We found that during assembly the results of \ac{asdf} are better compared to \ac{gbot} (Screw Clamp).  Due to the improvement of pose refinement, \ac{asdf} performs better than \ac{yolo}v8Pose (Screw Clamp, Nano Vise). }    
        \label{fig:gbot-images}
\end{figure}

\subsection{GBOT Results}

Existing approaches face the limitation of being either used for state detection combined with 2D object detection~\cite{stanescu_state-aware_2023} or are limited to non accessible datasets~\cite{murray_equipment_2024,su_deep_2019}. However, in terms of 6D pose estimation and 6D object tracking some approaches can handle 6D pose estimation for assembled parts. We focus on multi-state assembly comparisons, therefore, we omit comparing with single state approaches such as Mb-ICG~\cite{stoiber_multi-body_2022}, which require a re-initialization per assembly step. Our comparisons builds upon the benchmark proposed by Li et al. \cite{li_gbot_2024}. 

Given that we utilize the same objects as Li et al. \cite{li_gbot_2024}, we assess our trained models from the \ac{asdf} dataset on the \ac{gbot} dataset without the need for retraining on their training set. We exclude the comparison with LiftPod since the screws and small connectors are excluded from their evaluation due to a category-level problem. As illustrated in \autoref{tab:gbot}, our approach demonstrates superior performance compared to \ac{gbot} and their fusion of \ac{gbot} with deep learning-based re-initialization of object tracking. Moreover, in comparison to state-of-the-art tracking methods, Li et al. \cite{li_gbot_2024} introduce a deep learning-based approach for 6D pose estimation, which shares the same base architecture as our method. The outcomes presented in \autoref{tab:gbot} shows that our approach, outperforms their deep learning baseline.

Furthermore, we conduct a visual comparison between our approach and \ac{gbot}, including their initialization network \ac{yolo}v8Pose, as depicted in \autoref{fig:gbot-images}. The qualitative and quantitative analysis reveals that \ac{asdf} outperforms their deep learning-based and hybrid approach. Particularly, in the assembly process, \ac{asdf} exhibits notable improvements compared to \ac{gbot}.

\section{Discussion}

\ac{asdf} shows more robust results compared to a pure deep learning-based approach on our \ac{asdf} and the \ac{gbot} dataset. Robust 6D pose estimation and assembly state detection allow adding visual overlays, e.g. see \autoref{fig:gbot-images} or even highlighting the current assembly state, see \autoref{fig:asdf_prediction_state_pose}. The more robust and correct the prediction is, the more reliably an \ac{ar} interface can display this information.

Our evaluation has shown that combining object pose prediction with assembly state detection can lead to better results in 6D position estimation. However, the best performing approach is not always the most runtime friendly. 

The assembly processes are complicated by occlusions and state changes. Previous work aimed at constant 6D position estimation of mounted objects using object tracking. However, they lack the additional state information~\cite{stoiber_multi-body_2022,li_gbot_2024}. In addition, these approaches would have to reinitialize the tracking process in the event of occlusion. \ac{gbot} proposes the use of re-initialization with a real-time capable architecture. As the comparison with \ac{gbot} shows, we can even outperform their re-initialization approach in terms of 6D position estimation performance. This makes our work valuable in real-world scenarios.

The state transition, i.e. the change from state A to B, is a frequent challenge not only in the detection of assembly states. However, it is generally a challenge in phase detection or action detection~\cite{demir_deep_2023}. A common approach in this regard, which is not considered in our evaluation, is to define a time frame between transitions, which is not considered in the evaluation. We did not do this as we were aiming for a real-world scenario where constant knowledge of the assembly state and object position is crucial. Therefore, we included these potentially error-prone parts in our dataset. 

\subsection{Limitations}

In terms of performance we aimed for the highest accuracy. However, for the runtime this adds some additional overhead. 

Moreover, the \ac{asdf} dataset features texture-less objects which do not propose the challenge of reflective materials such as medical instruments. Nevertheless, texture less objects propose a challenge.

The current approach focuses on reliable pose estimations and assembly state detection using a single camera and for the dataset on reproducible items. To address occlusion a multi-camera approach or even fusing static and dynamic camera input could provide additional information for the estimation.

\section{Conclusion}

We present \ac{asdf} a deep learning-based approach utilizing late fusion for assembly state detection and 6D pose estimation. \ac{asdf} can track objects from assembly state zero (unassembled) until the full assembly is completed. This enables smart guidance and can be used in the medical or industrial context. 
On our assembly dataset, we outperform the deep learning-based without a fusion step. Moreover, \ac{asdf} has demonstrated that awareness of the assembly state leads to an improved performance compared to the state-of-the-art in 6D pose estimation on the \ac{gbot} dataset. The scenes in this dataset present various challenges, and \ac{asdf} has shown a huge improvement compared to \ac{yolo}v8Pose and outperformed all object tracking-based approaches.

In conclusion, our approach and dataset represent a promising step towards developing comparable backends for smart \ac{ar} guidance in assembly processes.

\acknowledgments{
This work is funded by the German Federal Ministry of Education and Research (BMBF) with grant number 16SV8973. 

We further thank d.hip for providing Hannah Schieber with a campus stipend.}

\bibliographystyle{abbrv-doi}

\bibliography{template}
\end{document}